\documentclass{article}

\usepackage[preprint, nonatbib]{neurips_2026}

\usepackage[utf8]{inputenc}
\usepackage[T1]{fontenc}
\usepackage{hyperref}
\usepackage{url}
\usepackage{booktabs}
\usepackage{amsmath}
\usepackage{amsfonts}
\usepackage{amssymb}
\usepackage{graphicx}
\usepackage{microtype}
\usepackage{placeins}
\usepackage{caption}
\title{Sharper Frames, Worse Rooms: Viewpoint Coverage Dominates Image Quality
in Feed-Forward 3D Reconstruction}

\author{%
  Alibek Dadajonov$^{*}$ \\
  Arbuz AI \\
  \texttt{alibek.dadajonov@arbuz.ai} \\
  \And
  Hikmatillo Umarhojiyev$^{*}$ \\
  Arbuz AI \\
  \texttt{hikmatillo.umarhojiyev@arbuz.ai} \\
}

\begin{document}
\maketitle

\begin{center}
\vspace{-1.5em}
$^{*}$Contributed Equally.
\end{center}

\begin{abstract}
Grounded spatial reasoning about indoor scenes rests on a geometric
reconstruction that the multimodal system does not itself produce.
Feed-forward multi-view reconstruction models operate under a fixed view
budget, making frame selection a critical upstream decision. We show that
under this setting the conventional preference for sharp frames can be
actively harmful. Across nine handheld room scans, selecting the sharpest 16
of 200 candidate frames is the worst strategy we test, reaching 48\%
room dimension error where uniform temporal sampling reaches 3\%. We trace
this failure to a coupling between image quality and camera motion: sharp
frames tend to occur during slow motion and therefore concentrate the
selection in a narrow angular region, leaving $192^\circ$ of the room
unobserved in the case we illustrate and over $300^\circ$ in the worst cells. A $2\times2$ factorial experiment that independently manipulates
viewpoint spread and per-frame sharpness shows that spread has roughly three
times the main effect of quality, and that blurry but well spread selections
outperform sharp but clustered selections in 8 of 9 captures. The advantage
also diminishes as the frame budget increases and angular coverage saturates,
consistent with this mechanism. Finally, we ask whether the failure is
detectable at inference: the reconstruction model's multi-view consistency
score increases with reconstruction error, while a vision-language model
given the same frames reports confidence unrelated to its own spatial
estimation error. Under small view budgets, viewpoint coverage is a more
important constraint than per-frame image quality, and neither uncertainty
signal available at inference reveals this upstream failure.
\end{abstract}

\section{Introduction}

A phone scanning a room produces on the order of 1{,}800 frames. The
reconstruction model consuming them accepts sixteen. Something has to choose,
and that choice fixes the geometry every downstream task inherits.

The literature on this choice is mature. Classical keyframe selection for
structure-from-motion maximises triangulation baseline, avoids degenerate
configurations, and rejects motion blur
\cite{ahmed2010,pollefeys2002,torr1998}. Later work adds learned blur
classifiers \cite{koschel2021}, blur metrics coupled to flight geometry
\cite{zhang2017}, and appearance change signals \cite{jha2025}. Across all of
it, per-frame image quality is something to maximise and blur is something to
remove. We found no prior work that evaluates this trade-off under a fixed
frame budget in this capture regime.

We ask that question in a setting where the answer turns out to matter. Our reconstruction
backbone is MapAnything \cite{keetha2025}, a feed-forward transformer that
ingests $N$ views jointly and regresses metric geometry. It computes no
correspondences, estimates no fundamental matrix, and runs no bundle
adjustment. These are substantially different assumptions from the pipelines
the classical criteria were designed for, which motivates testing whether the
criteria transfer. Note also that the configurations those criteria explicitly
reject, namely rotation about the camera centre and coplanar structure,
describe a handheld sweep of a room with flat walls.

Our hypothesis is that under a fixed frame budget and rotation dominant
handheld capture, reconstruction error is governed more by viewpoint coverage
than by per-frame image quality. The hypothesis was motivated by observing that sharpness
ranked selection failed badly, so \S4.1 and \S4.2 are confirmatory rather than
predictive; \S4.3 tests a prediction made before running it, that the
advantage of spread should shrink once any reasonable selection covers the
scene.

The reliability angle is what makes this a deployment problem rather than a
tuning problem. A pipeline that silently loses viewpoint coverage does not
merely become less accurate. It produces geometry that is confidently wrong,
and the signal a practitioner would reach for to detect that, the model's own
multi-view consistency score, moves in the wrong direction. Downstream
multimodal reasoning about the scene inherits the corrupted geometry with no
indication that anything failed.

The budget is not a matter of taste either. The deployed system we study
serves a reconstruction graph exported at a fixed sixteen views, so selection
is a hard cardinality constraint imposed by the architecture.

Our contributions:

\begin{itemize}
\itemsep2pt
\item Sharpness ranked selection is the worst performing strategy we test,
worse than random at $K{=}16$ and worse at covering the scene at every budget,
and we identify the mechanism: sharp frames
co-occur with slow camera motion, so ranking by sharpness ranks by angular
redundancy.
\item A $2\times2$ factorial isolates that mechanism, manipulating spread and
quality independently. Both matter; spread matters about three times more.
\item The effect has a predictable boundary. Sweeping $K$ from 8 to 48, both
the error gap and the catastrophic failure rate collapse where angular
coverage saturates.
\item Neither confidence signal available to a practitioner detects the
failure. The model's multi-view consistency score increases with error in
every capture we test, and a VLM's self-reported confidence carries no usable
information about its own error.
\end{itemize}

\section{Related Work}

\paragraph{Triangulation-era criteria.}
Ahmed et al.~\cite{ahmed2010} select keyframes using a correspondence ratio as
a baseline proxy, Torr's GRIC \cite{torr1998} to choose between homography and
fundamental matrix models, and a point-to-epipolar-line cost, building on
earlier correspondence-ratio and reprojection-error criteria
\cite{seo2003,seo2008} and on hand-held camera modelling \cite{pollefeys2002}. Their stated
motivation is triangulation accuracy: uncertainty grows as inter-view distance
shrinks, so selection should maximise baseline. They also disqualify
candidates under \emph{motion degeneracy} (rotation about the camera centre,
leaving epipolar geometry undefined) and \emph{structure degeneracy} (coplanar
points, leaving $F$ undetermined). Koschel et al.~\cite{koschel2021} note
independently that this discards frames carrying real information and presumes
a recording strategy without camera rotation about its axis. That the excluded
regime is ours is not incidental: Rashidi et al.~\cite{rashidi2012} construct
validation sequences by deliberately filming planar walls and by rotating the
camera without translation, precisely so that their selector can be shown to
remove such frames.

\paragraph{Quality-based criteria.}
Koschel et al.~\cite{koschel2021} train a random forest on inertial data to
reject motion-blurred and redundant frames, validated by reprojection RMSE
under Meshroom. Zhang et al.~\cite{zhang2017} apply the Crete blur metric
\cite{crete2007} to drone footage. Rashidi et al.~\cite{rashidi2012} add
blurring restrictions to prior selection methods. In each case blur rejection
is validated against a geometric proxy, never at fixed cardinality, and never
against a coverage-matched control. Koschel et al.\ come closest to posing the
question: their Figure 1 contrasts a sequence with good image quality but no
change in content against one with viewpoint variation but motion blur. They
resolve it toward quality and leave overlap-based selection to future work.

\paragraph{Coverage-based selection.}
Coverage objectives are established prior art and we claim no novelty for the
selection rule. Wang et al.~\cite{wang2014} optimise a global viewing angle
coverage objective at fixed cardinality via transductive experimental design;
the angular bin selector we use is their comparison heuristic, dividing
$360^\circ$ into 36 bins and taking the best frame per bin. Zhang
et al.~\cite{zhang2017} enforce a pairwise overlap rate derived from altitude
and field of view. Neither work puts coverage and quality in competition:
Wang et al.\ consider no image quality at all, and Zhang et al.\ fix geometric
spacing first, treating blur as a local refinement within it.

These settings also guarantee angular coverage by construction: gimbal
stabilised nadir flight at constant velocity \cite{zhang2017}, a rig mounted
$360^\circ$ camera \cite{koschel2021}, or many contributors circling a landmark
\cite{wang2014}. Each reports coverage aware selection performing close to
uniform sampling. A single handheld interior sweep is different: our unwrapped
yaw trajectories span $386^\circ$ to $1083^\circ$ on the eight full sweeps and
are non-monotonic, so
temporal uniformity decouples from angular coverage.

\paragraph{Feed-forward multi-view reconstruction.}
The feed-forward family descends from DUSt3R \cite{dust3r2024}, which
regresses pointmaps from image pairs, through unified transformers that ingest
many views in a single pass \cite{vggt2025}, streaming variants with
persistent state \cite{cut3r2025}, and MapAnything \cite{keetha2025}, which
regresses factored metric geometry from $N$ views jointly; see
\cite{ffsurvey2025} for a survey. Most of these anchor their output to a designated reference view, an inductive
bias that Wang et al.~\cite{pi3} identify as a source of instability when the
reference is suboptimal; their $\pi^3$ adopts a fully permutation-equivariant
architecture specifically to remove that dependence and the input-ordering
sensitivity it creates. Jha et al.~\cite{jha2025} study adaptive keyframe
selection for Spann3r and CUT3R, the same model family, using photometric and SSIM change
signals; they assume RGB-D input, use threshold rather than fixed cardinality
selection, evaluate on Chamfer distance, and include no coverage baseline.
Frame selection as cardinality constrained submodular maximisation has been
formalised for video language models \cite{mdp3}; our bin coverage objective is
a coverage function and inherits the standard $(1-1/e)$ greedy guarantee
\cite{nemhauser1978}, which we note for rigour rather than as a contribution.

\paragraph{Uncertainty in learned reconstruction.}
Learned depth predictors are known to be overconfident in difficult conditions
\cite{evidmtl2025}, a problem that motivates evidential and post-hoc
uncertainty methods, and post-hoc uncertainty estimation for monocular depth is
an active area \cite{gradunc2025}. Uncertainty in learned priors has also been treated directly: MP-SfM
\cite{mpsfm2025} augments classical structure-from-motion with monocular depth
and normal priors and propagates their uncertainty, reporting robustness to
errors in those priors across models. Most relevant here, UfM \cite{ufm2026} argues
that single image uncertainty misses disagreement between predictions across
views of the same region, and proposes multi-view disagreement as a more
informative signal. Our results qualify that proposal rather than contradict
it. Multi-view agreement is informative about many error sources, but it is
anti-informative about the one introduced by frame selection, because a
selection that loses angular coverage is also a selection whose views are
mutually redundant. We are not aware of prior work reporting that a multi-view
consistency score increases with reconstruction error.

\section{Method}

\subsection{Data and features}

Nine handheld phone videos of single rooms in residential and office
interiors, 30\,fps, one to two minutes each. From each we extract a candidate
pool of 200 frames at even temporal spacing and compute two per-frame features.

\textbf{Sharpness} is the variance of the Laplacian, percentile normalised
within a capture.

\textbf{Yaw} is integrated from optical flow using a nominal horizontal field
of view. Since this is the input to every coverage-based strategy we validate
it against camera orientations recovered from a dense reconstruction: eight of
nine captures give Pearson $r \geq 0.70$ (median 0.87, four captures at
$r \geq 0.98$). Optical flow systematically overestimates total rotation
because translation contaminates the flow field, so we rely on the correlation
rather than the absolute scale. One capture fails the check at $r = 0.154$ and
we treat it separately throughout (\S\ref{sec:limits}).

\subsection{Reconstruction and metric extraction}

Reconstruction uses MapAnything with mixed precision inference and edge
masking. Dimensions are extracted by pooling masked world points, recovering
gravity as the mean camera up-vector, aligning yaw by minimum area rectangle,
and taking 5 to 95 percentile extents along the two horizontal axes.

Two checks on this procedure. The gravity estimate agrees with an independent
scene layeredness estimate to $6.9^\circ$, and the resulting ceiling heights
(2.24 to 2.47\,m) fall in the expected range for residential interiors. The
pipeline is deterministic: full reruns reproduce every reported error and
failure rate to three decimals. We use the Apache-2.0 licensed
\texttt{map-anything-apache} checkpoint. All reconstruction ran on a single
NVIDIA GB10 with unified memory; one reconstruction takes about 2\,s at
$K{=}16$ and 5\,s at $K{=}32$, and the full set of experiments reported here
amounts to roughly 4 GPU-hours.

\subsection{Reference and its noise floor}
\label{sec:ref}

We have no tape measure ground truth. Our reference is a dense reconstruction
over all 200 pool frames, so reported errors measure disagreement with a dense
reconstruction rather than metric accuracy. To bound how much of that
disagreement is reference noise, we reconstruct from two disjoint halves of
each pool (even and odd indices, 100 frames each, both densely covering the
room) and measure how much the two dense estimates differ.

Mean disagreement is 3.3\% across nine captures, and 1.6\% excluding the
capture that also fails yaw validation, with a range of 0.3 to 5.2\%. This
bounds the reference's \emph{stability}, not its accuracy: two dense
reconstructions agree closely with each other while both being wrong in the
same way. We therefore hand measured the three rooms we could physically
access. The dense reference differs from the tape by 8.1\%, 11.4\% and 11.8\%,
a mean absolute error of 10\% with no consistent direction and roughly an order
of magnitude larger than the stability figure. These same three captures have
the best self-consistency in the set, so internal agreement between dense
reconstructions does not predict agreement with reality. Room shape is
nonetheless recovered: the square room reconstructs as square.
Table~\ref{tab:tape} gives the per-room figures.

That figure sets what the reported errors can and cannot support. Our
best-performing cells sit at 11.0\% and 15.4\% error against the reference,
which is the same magnitude as the reference's own error, so we make no claim
that those reconstructions are accurate in absolute terms. The clustered cells
sit at 32.4\% and 43.5\%, two to three times the reference uncertainty, and the
spread main effect of 0.247 (95\% CI $[0.155, 0.340]$) lies well outside it.
The comparison between selection strategies is therefore supported; the
absolute accuracy of any single reconstruction is not.

\begin{table}[t]
\centering
\small
\begin{tabular}{lccc}
\toprule
capture & measured (m) & dense reference (m) & relative error \\
\midrule
bedroom & $3.39 \times 3.39$ & $3.94 \times 3.63$ & 11.8\% \\
work\_office     & $3.90 \times 2.62$ & $4.22 \times 2.83$ & 8.1\% \\
home\_office    & $4.72 \times 3.72$ & $4.04 \times 3.41$ & 11.4\% \\
\bottomrule
\end{tabular}
\vspace{5mm}
\caption{Hand measured dimensions against the dense 200-frame reference, for
the three captures we could physically access. Mean absolute error is 10\%
with no consistent direction: the reference overestimates both dimensions of
bedroom and work\_office and underestimates both dimensions of
home\_office. Room shape is nonetheless recovered, with the square room
reconstructed as square. Note also that these three captures have the best
self-consistency in the set (0.3 to 1.1\%, Section~\ref{sec:ref}) while showing
8 to 12\% accuracy error, so internal agreement between dense reconstructions
does not predict agreement with reality.}
\label{tab:tape}
\end{table}

\subsection{Selection strategies}

All strategies map a feature table and budget $K$ to $K$ indices.
\textbf{Uniform} takes even temporal spacing, the deployed baseline.
\textbf{Random} samples without replacement. \textbf{Sharpness} takes the
top-$K$ by Laplacian variance. \textbf{Coverage} performs greedy
marginal gain selection over 36 yaw bins of $10^\circ$, following
\cite{wang2014}; \textbf{Coverage-FPS} substitutes farthest point sampling in
yaw; \textbf{Coverage-$q$} restricts the greedy pool to frames above the
$q$-th sharpness percentile.

The four factorial cells cross \emph{spread} (greedy bin coverage) against
\emph{clustered} (a contiguous run), and \emph{high} against \emph{low}
quality (above or below median sharpness). The informative cell is
spread-low: blurry frames that nevertheless see the whole room.

\paragraph{Effects and experimental unit.} Main effects are capture level
differences of marginal means, signed so that positive values mean clustering
or low sharpness increases error; the interaction is the difference of
differences. Each cell contains 81 reconstructions, but these are nine captures
$\times$ three seeds $\times$ three view orderings, and seeds and orderings are
repeated measures within a capture. All inferential statistics therefore
collapse to per-capture cell means and test across the nine captures. Cell
means and standard deviations are descriptive summaries over all
reconstructions; $p$-values and confidence intervals use $n{=}9$.

\subsection{Metric}

We report two quantities and never collapse them. \textbf{Degeneracy rate} is
the fraction of reconstructions failing a geometric test fixed before any
reconstruction was run: a
minor dimension below 1.2\,m or an aspect ratio above 4:1, neither of which
describes a habitable room. \textbf{Relative dimension error} is the mean absolute
relative error on the two principal horizontal extents, computed over
\emph{all} reconstructions including degenerate ones. Conditioning on success
discards the worst runs; in an earlier version of this analysis it understated
the worst cell's penalty by 8 points.

\subsection{View ordering}

Feed-forward multi-view models anchor their output to a designated reference
view and are therefore sensitive to input ordering, a dependence $\pi^3$
\cite{pi3} was designed to remove. Reordering the identical sixteen frames changes the
recovered long dimension by 1.5 to 13.7\% per capture on the validated captures
(median $\approx$9\%), yet the between-strategy error spread exceeds the
ordering variance in all nine captures. We fix temporal order for the strategy
comparison and additionally marginalise the factorial over three random
orderings per selection, so this variance sits inside the reported error bars.

\section{Results}

\begin{figure}[htbp]
\centering
\includegraphics[width=\linewidth]{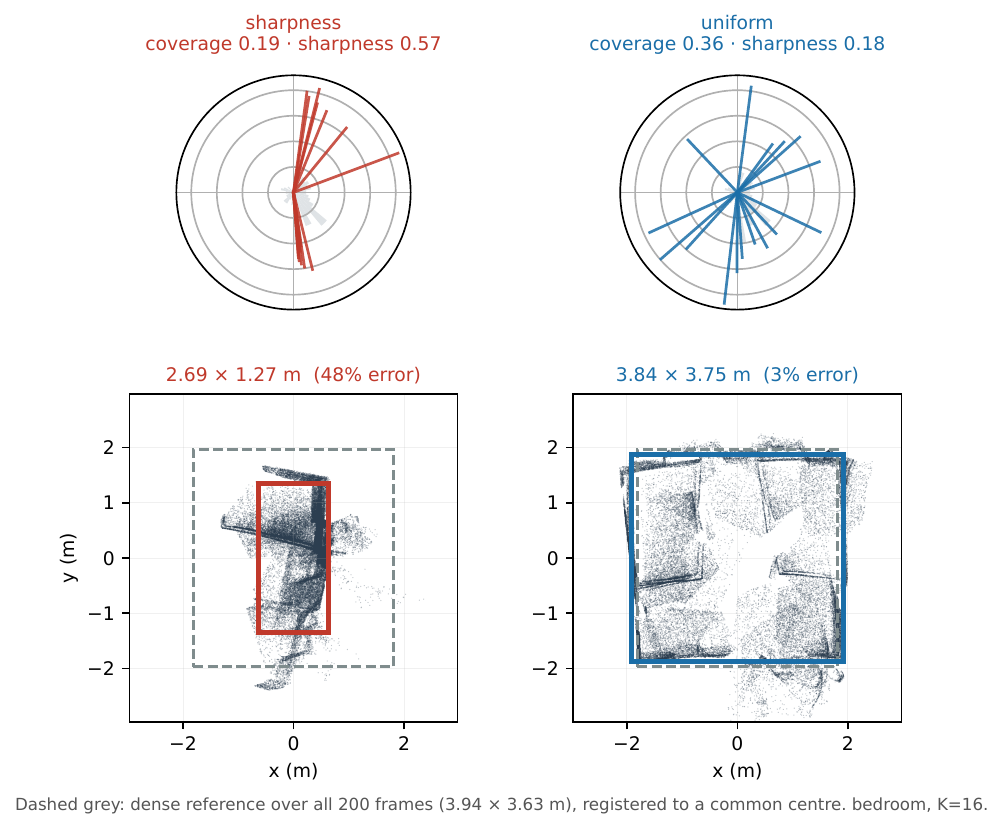}
\caption{Sharpness ranked selection against uniform sampling on one capture
($K{=}16$). Top: viewing directions of selected frames (colour) over the yaw
distribution of all 200 candidates (grey). Bottom: reconstructed point cloud
from above, with the fitted rectangle (solid) and the dense 200-frame
reference (dashed), registered to a common centre. Rectangles use 5 to 95
percentile extents, so sparse peripheral points fall outside them.}
\label{fig:main}
\end{figure}

\subsection{Sharpness ranking is the worst strategy tested}

At $K = 16$ across nine captures and three seeds, sharpness ranked selection
gives the highest error of the seven named strategies (0.278), against 0.063
for uniform sampling and 0.114 for random, reaching 48\% on the worst capture
(Fig.~\ref{fig:main}). That selection achieves 0.194 angular coverage with a
largest unobserved wedge of 3.35\,rad, so $192^\circ$ of the room is never
seen. The deficit is not a small budget artefact: even at $K = 48$, sharpness
reaches only 0.562 angular coverage where random sampling reaches 0.704.
Ranking by sharpness is, in this capture regime, ranking by angular
redundancy. This comparison fixes temporal view order rather than marginalising
over it, so the large gaps hold but fine distinctions among the well-performing
strategies sit within the ordering variance of \S3.6 and we draw no conclusion
from them.

\subsection{The factorial: spread dominates, and the factors interact}

\begin{table}[t]
\centering
\small
\begin{tabular}{lcccc}
\toprule
& \multicolumn{2}{c}{Relative error} & \multicolumn{2}{c}{Degenerate} \\
\cmidrule(lr){2-3}\cmidrule(lr){4-5}
& high quality & low quality & high quality & low quality \\
\midrule
spread     & $0.110 \pm 0.077$ & $0.154 \pm 0.131$ & 0.000 & 0.025 \\
clustered  & $0.324 \pm 0.152$ & $0.435 \pm 0.166$ & 0.086 & 0.395 \\
\bottomrule
\end{tabular}
\vspace{5mm}
\caption{$2\times2$ factorial at $K{=}16$, marginalised over view ordering.
Means and standard deviations over 81 reconstructions per cell from nine
captures; inference treats the capture as the unit ($n{=}9$).}
\label{tab:factorial}
\end{table}

\begin{figure}[t]
\centering
\includegraphics[width=\linewidth]{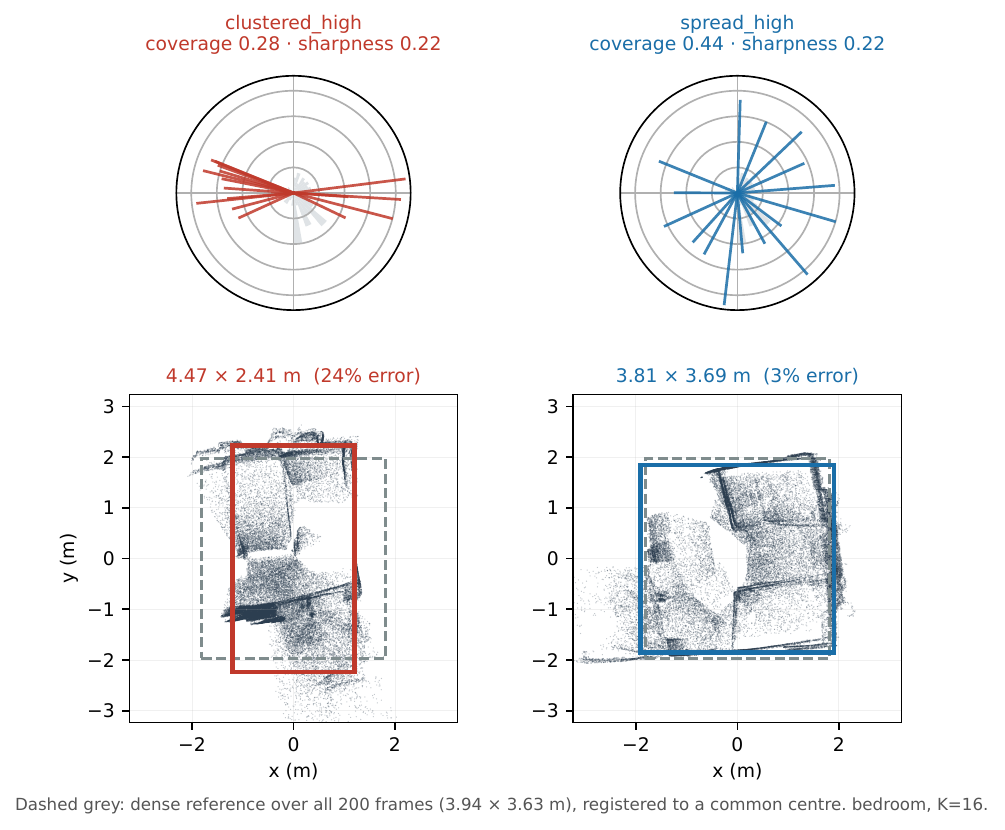}
\caption{The factorial contrast on one capture: clustered-high against
spread-high, identical mean sharpness (0.22), differing only in spread. Same
conventions as Figure~\ref{fig:main}.}
\label{fig:factorial}
\end{figure}

Table~\ref{tab:factorial} gives the result; Figure~\ref{fig:factorial}
visualises the same contrast on a single capture.
Paired across the nine captures, the spread main effect is $+0.247$
(95\% CI $[+0.155, +0.340]$, $p = 0.0003$, positive in 9/9) and the quality
main effect $+0.077$ (95\% CI $[+0.018, +0.136]$, $p = 0.017$, 8/9). Two things
beyond those numbers.

First, spread-low beats clustered-high by a factor of two, in 8 of 9 captures
(paired difference $+0.170$, paired $t$-test $p = 0.009$; Wilcoxon signed rank
$p = 0.012$). Because spread and quality are
manipulated independently, this directly compares reconstruction's sensitivity
to the two factors under a fixed budget: if per-frame quality were the binding
constraint the ordering would reverse. The single exception is the capture that fails yaw
validation (\S\ref{sec:limits}); on the eight validated captures the result is
unanimous.

One caveat on the ratio: our spread manipulation is extreme while quality is a
median split, so the factor of three reflects the contrast widths we chose
rather than an intrinsic property of the factors. The direct cell comparison
above does not depend on it.

Second, the cell means suggest the two factors interact. Moving from high to
low quality costs 0.044 error under spread and 0.111 under clustering, and
takes degeneracy from 0\% to 2.5\% versus 8.6\% to 39.5\%. The direction is
consistent in 8 of 9 captures, but the interaction term is not statistically
supported at the capture level ($+0.067$, 95\% CI $[-0.064, +0.199]$,
$p = 0.27$). We report it as a suggestive pattern that a larger capture set
would be needed to establish.

Degenerate reconstructions are wrong in a different way rather than noisy:
typical failures return rooms 0.59 to 0.89\,m wide with unobserved wedges of
4.2 to 5.7\,rad, a slab rather than a room.

\subsection{Budget sweep: the mechanism predicts its own boundary}

If the mechanism is coverage, the advantage should shrink once coverage stops
being scarce. With 36 bins, full bin coverage requires at least 36 selected
frames. Table~\ref{tab:budget} sweeps $K$ over four
full sweep captures and
shows two independently measured quantities collapsing at that point: the
spread advantage declines gradually from 0.262 to 0.190 through $K{=}32$ and
then halves to 0.105 at $K{=}48$, where our spread selections first reach
coverage 1.000, and the degeneracy rate of the worst cell falls monotonically
from 50\% to 0\% on the same schedule. Spread error is flat over the last two budgets (0.078, 0.087), so the
closing gap comes from clustered selections improving. This
also locates the finding, since the deployed system runs at $K{=}16$, near the
maximum of the effect.

\begin{table}[htbp]
\centering
\small
\begin{tabular}{lcccc}
\toprule
$K$ & 8 & 16 & 32 & 48 \\
\midrule
clustered error            & 0.421 & 0.316 & 0.268 & 0.192 \\
spread error               & 0.160 & 0.104 & 0.078 & 0.087 \\
\textbf{gap}               & \textbf{0.262} & \textbf{0.212} & \textbf{0.190} & \textbf{0.105} \\
\midrule
coverage achieved (spread) & 0.222 & 0.444 & 0.889 & 1.000 \\
degeneracy, clustered-low  & 0.500 & 0.417 & 0.167 & 0.000 \\
\bottomrule
\end{tabular}
\vspace{5mm}
\caption{Budget sweep over the four full sweep captures available when it was
run, three seeds each, so twelve reconstructions per cell. $K{=}12$ and
$K{=}24$ are omitted for space and interpolate smoothly.}
\label{tab:budget}
\end{table}

\subsection{Confidence increases with error}

Cell means are given in Table~\ref{tab:conf}. The
signal available at inference time is a multi-view consistency score:
the agreement between per-view depth predictions where they overlap.
Practitioners naturally read it as a reliability estimate. It \emph{increases}
with error. Within each capture, the Spearman
correlation between confidence and error is positive in 9 of 9 captures, mean
$\rho = +0.529$, individually significant in 8 (sign test across captures
$p = 0.004$; Wilcoxon $p = 0.004$). Pooling reconstructions across captures
weakens the apparent relationship rather than strengthening it, because
between-capture difficulty varies more than confidence does. Degenerate
reconstructions carry a 41\% higher mean consistency score than valid ones
from the same capture ($0.710$ vs $0.505$, paired across the eight captures
containing both, $p = 0.003$, 7/8 captures).

This is not a defect in the score so much as a mismatch between what it
measures and what a practitioner would use it for. Clustered views observe the
same surfaces from similar angles, so their depth predictions agree, and
agreement is exactly what the score reports. Consistency is not correctness.
Under frame selection the two move in opposite directions, because the
selections that lose coverage are the same selections that gain redundancy.

For a deployed system the implication is uncomfortable. A pipeline that selects
frames by sharpness, a defensible looking choice with substantial literature
behind it, will produce systematically wrong geometry while reporting elevated
confidence.

\begin{table}[t]
\centering
\small
\begin{tabular}{lcc}
\toprule
& relative error & multi-view confidence \\
\midrule
spread, high quality    & 0.110 & 0.444 \\
spread, low quality     & 0.154 & 0.363 \\
clustered, high quality & 0.324 & 0.669 \\
clustered, low quality  & 0.435 & 0.674 \\
\bottomrule
\end{tabular}
\vspace{5mm}
\caption{Multi-view consistency score against error, means over 81
reconstructions per cell. The ordering is inverted throughout.}
\label{tab:conf}
\end{table}

\section{Is an image-only VLM affected by the same selection?}
\label{sec:vlm}

We asked whether the frame selection that breaks the geometry also degrades a
VLM's own spatial estimate. It does not, and the VLM offers no compensating
reliability signal: it cannot cross-check a corrupted reconstruction, because
its errors are unrelated to the geometry's and its reported confidence is
uninformative about its own.

We gave a vision-language model (\texttt{claude-sonnet-5}, queried August 2026)
the same selected frames and asked it to
estimate the room's floor dimensions and its own confidence, over
all nine captures, seven strategies and three seeds (189 queries).

\paragraph{Frame selection barely affects the VLM.}
Strategy means span 0.184 to 0.270 with standard deviations of 0.07 to 0.17, so
the between-strategy spread is smaller than the within-strategy noise. The
$2\times2$ ordering is even mildly inverted relative to the geometric result
(clustered 0.213, spread 0.248), and sharpness-ranked selection, the worst
strategy for reconstruction, gives the lowest VLM error of the seven.

\begin{figure}[t]
\centering
\includegraphics[width=\linewidth]{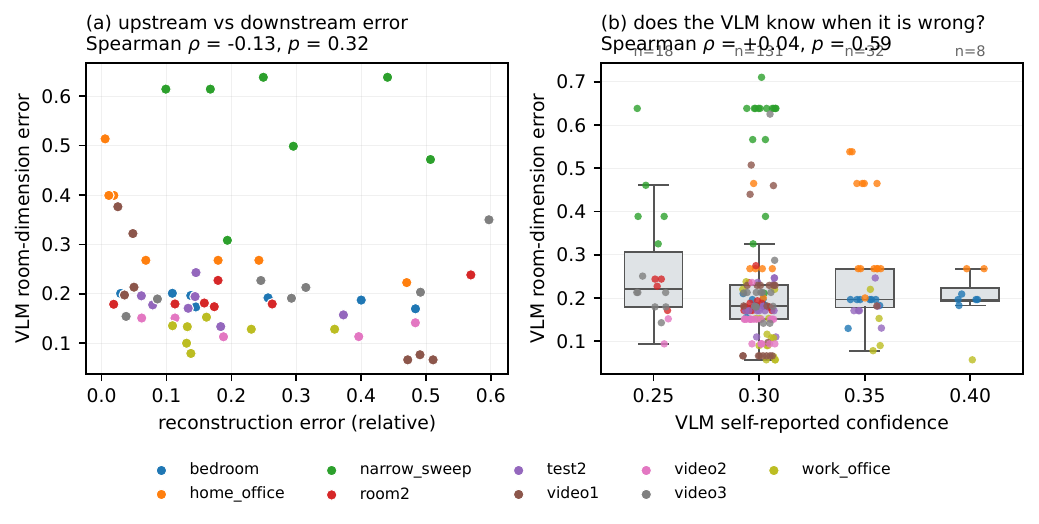}
\caption{Neither downstream signal tracks the upstream failure. (a) Per
capture and strategy, reconstruction error against the VLM's room-dimension
error: the selections that break the geometry are not the ones the VLM gets
wrong. (b) Error distribution at each of the four confidence values the
model actually reports, with individual queries jittered over the boxes. Of 189
queries, 131 return a confidence of exactly 0.30 and the 0.40 bin contains only
eight, so the outer bins should not be over-read. Colour denotes capture.
Section~\ref{app:prompt} gives the query text.}
\label{fig:vlm}
\end{figure}

\paragraph{VLM error is capture dominated.}
Per-capture mean error ranges from 0.122 to 0.541, while three captures have
within-capture standard deviations of 0.018, 0.031 and 0.036 across all seven
strategies. On one capture the VLM returns essentially the same estimate
whatever frames it is shown ($0.188 \pm 0.018$). Which room is being estimated
explains substantially more variance than which frames are selected. Geometric
and VLM error are also uncorrelated across the 63 matched capture-strategy
cells (Spearman $-0.127$, $p = 0.32$; Figure~\ref{fig:vlm}a).

One possible explanation is that the two systems solve the task by different
routes. MapAnything aggregates geometry across views and therefore needs the
views to cover the room, whereas a VLM may rely more heavily on visual priors
such as furniture scale, door and ceiling heights, and perspective, which one
adequate view can supply. We did not measure what the model attends to, so
this remains a hypothesis.

\paragraph{VLM confidence provides no reliable error signal.}
Across all 189 queries the model emits only four distinct confidence values,
0.25, 0.30, 0.35 and 0.40, so per-strategy means sit in a narrow band from
0.287 to 0.320 while error ranges from 0.184 to 0.270. Confidence and error are
uncorrelated (Spearman $+0.039$, $p = 0.59$), and the error distribution is
essentially unchanged across the four reported levels
(Figure~\ref{fig:vlm}b). The signal is not merely miscalibrated; it barely
varies.
The prompt asked for a checkable quantity, the probability that the estimate
falls within 15\% of the true value (Appendix~\ref{app:prompt}), so this is
not a failure to interpret a vague question.

We note also that a pooled correlation between angular coverage and VLM error
that appeared strong on a five-capture subset ($\rho = -0.37$) falls to
$\rho = -0.125$ ($p = 0.09$) on the full set, consistent with it having been
between-capture variance rather than a selection effect.

Taken with \S4.4, the deployment picture is that neither signal in the stack
was usable in our experiments. The reconstruction model's consistency score is
highest where accuracy is lowest. The VLM's stated confidence carries no usable
information about its own error, and its errors are unrelated to the
geometry's, so in this task it could not act as a check on a corrupted
reconstruction either.

\section{Limitations}
\label{sec:limits}

\textbf{No absolute ground truth.} Errors measure disagreement with a dense
reconstruction from the same model, not metric accuracy; Section~\ref{sec:ref}
bounds its noise at 3.3\% and its accuracy at 10\%. Two findings do not depend
on the reference: degeneracy rates rest on physically impossible dimensions,
and the confidence results are internal to the models. Uniform sampling is also
structurally advantaged as the $K$-frame subsample most resembling the
reference, so we make no claim about it being optimal.

\textbf{Nine single room captures, three with physical ground truth.} All are
residential or office interiors. The remaining six were not accessible for
measurement. The effect is consistent across them, but the
sample is small and we do not test unusual room shapes.

\textbf{One capture fails yaw validation and is anomalous throughout.} A
narrow sweep capture ($80^\circ$ unwrapped) gives $r = 0.154$ against
reconstructed camera orientations, 16.4\% reference disagreement against a
1.6\% mean elsewhere, 0.541 VLM error against 0.122 to 0.334 elsewhere, the
largest ordering variance (18.7\%), and is the only capture in our set where
spread-low loses to clustered-high. We report it in aggregates but draw no boundary
condition from it. Characterising behaviour under narrow sweeps needs captures
whose yaw estimate validates.

\textbf{Scope of the experiments.} The strategy comparison, factorial,
consistency score, reference, yaw and downstream VLM analyses use all nine
captures; the budget sweep uses four. We also use a single sharpness measure,
Laplacian variance, whereas the works we contrast with use a perceptual blur
metric \cite{crete2007} or DFT magnitude \cite{koschel2021}; the factorial
partitions at the median and is largely insensitive to this, but the top-$K$
sharpness strategy could rank frames differently under a directional measure.

\textbf{Implementation gap with the deployed system.} Our experiments use the
PyTorch reference implementation, because the deployed ONNX export is fixed at
16 views and pads shorter inputs by repeating the last frame, which would
confound any comparison at $K \neq 16$. On one capture the deployed service
returns $3.63 \times 3.60$\,m where our dense reference returns
$3.94 \times 3.63$\,m, about 8\% apart. All our claims are relative
comparisons within one implementation, so this does not affect them, but our
absolute values should not be read as the deployed system's.

\textbf{Propagation through the reconstruction is untested.} Section~5 gives
the VLM the same frames, not the reconstruction, so it tests whether an
image-only VLM shares the selection sensitivity, not whether corrupted geometry
poisons a system reasoning over it.

\textbf{One reconstruction model.} Whether the finding transfers to other
feed-forward multi-view backbones, particularly permutation-equivariant ones,
is open. It would also test whether the order sensitivity we measure is
intrinsic to the architecture class.

\section{Conclusion}

Frame selection for feed-forward multi-view reconstruction is not the same
problem as keyframe selection for structure-from-motion, and the inherited
heuristic of keeping the sharp frames is harmful when the budget is small. The
cause is not subtle: sharp frames cluster where the camera moved slowly, and a
room you did not look at cannot be reconstructed. Under a fixed budget, the estimated main
effect of viewpoint coverage is roughly three times that of per-frame image
quality, and blurry frames that see the whole room beat sharp frames that see one
wall in 8 of 9 captures.

For deployment the reliability consequence matters more than the accuracy one.
An upstream heuristic with substantial literature behind it can silently
corrupt the geometry a multimodal system reasons over, and neither uncertainty
signal available at inference gives warning: the reconstruction model's
consistency score is highest exactly when it is most wrong, and in our
experiment a VLM's spatial estimates were insensitive to the selection that
broke the geometry, with confidence that carried no error signal. In a pipeline
where semantic reasoning is grounded in reconstructed geometry, that is a
failure mode with no detector. The practical guidance follows: when the budget
is small relative to the angular extent of the scene, select for coverage and
let sharpness follow, and do not treat either confidence signal as a proxy for
correctness.

\clearpage
\appendix

\section{VLM query details}
\label{app:prompt}

The downstream experiment in Section~5 queries
\texttt{claude-sonnet-5} through its public API. Each query contains the $K{=}16$
selected frames in temporal order, resized so the long side is 768\,px and
re-encoded as JPEG at quality 85, followed by the text below. Output is capped
at 300 tokens. Responses are parsed as JSON; a response that fails to parse is
retried up to four times with exponential backoff and recorded as a failure if
it never parses.

\begin{quote}
\ttfamily\small\raggedright
These are 16 frames sampled from a single continuous video of ONE room, in
temporal order.\\[4pt]
Estimate the room's floor dimensions.\\[4pt]
Reply with ONLY a JSON object, no other text:\\
\{"length\_m": <number>, "width\_m": <number>, "ceiling\_m": <number>,
"confidence": <0-1>, "reasoning": "<one sentence>"\}\\[4pt]
length\_m is the longer horizontal dimension, width\_m the shorter, both in
metres. confidence is how sure you are that your estimate is within 15\% of
the true value.
\end{quote}

We report \texttt{length\_m} and \texttt{width\_m} sorted so the larger is
first, matching the convention used for the geometric metric, and
\texttt{confidence} as the self-reported reliability estimate. The
\texttt{ceiling\_m} and \texttt{reasoning} fields were recorded but are not
analysed here.

Note that the confidence field asks for a specific, checkable quantity, namely
the probability that the estimate falls within 15\% of the true value, rather
than a vague self-assessment. The reported values cluster between 0.29 and
0.33 across every condition, which is part of why they carry so little
information about the actual error.

\end{document}